\documentclass[preprint,12pt]{elsarticle}

\usepackage[utf8]{inputenc}
\usepackage[T1]{fontenc}
\usepackage{amsmath,amsfonts,amssymb}
\usepackage{graphicx}
\usepackage{hyperref}
\usepackage{enumitem}
\usepackage{natbib}
\usepackage{amsthm} 

\usepackage{algorithm}
\usepackage{algpseudocode}
\usepackage{booktabs}
\usepackage{multirow}
\usepackage{array}
\usepackage{soul}
\usepackage{subcaption} 
\usepackage[dvipsnames]{xcolor}
\usepackage{pifont}

\begin{document}

\begin{frontmatter}
\title{VGG Induced Deep Hand Sign Language Detection}
\author[1]{Subham Sharma}
\ead{sbhm.sbhm@gmail.com, subham@dxops.ai}
\author[1,2]{Sharmila Subudhi\corref{cor1}}
\ead{sharmilasubudhi@ieee.org}
\affiliation[1]{organization={DXOps}, city={Bangalore}, country={India}}
\affiliation[2]{organization={Dept. of Computer Science, Maharaja Sriram Chandra Bhanja Deo University}, city={Baripada}, postcode={757003}, state={Odisha}, country={India}}
\cortext[cor1]{Corresponding author}
\begin{abstract}
Hand gesture recognition is an important aspect of human-computer interaction. It forms the basis of sign language for the visually impaired people. This work proposes a novel hand gesture recognizing system for the differently-abled persons. The model uses a convolutional neural network, known as VGG-16 net, for building a trained model on a widely used image dataset by employing Python and Keras libraries. Furthermore, the result is validated by the NUS dataset, consisting of 10 classes of hand gestures, fed to the model as the validation set. Afterwards, a testing dataset of 10 classes is built by employing Google’s open source Application Programming Interface (API) that captures different gestures of human hand and the efficacy is then measured by carrying out experiments. The experimental results show that by combining a transfer learning mechanism together with the image data augmentation, the VGG-16 net produced around 98\% accuracy.
\end{abstract}

\end{frontmatter}

\section{Introduction}\label{sec:introduction}
Human computer interaction is the core of all intelligent systems we see around us. Common types of human computer interaction are using devices like, keyboard and mouse, etc. A natural way of human interaction is the usage of hand gestures  \cite{sm2000}. These gestures give a better understanding while a speech is conveyed. The gestures are also useful in sign language interaction when the speech is absent. An advantage of using hand gestures for human computer interaction over others is the speed and ease. The messages can be conveyed faster. Integrating human gestures recognition system with human computer interaction serves to be a significant step in human-human and human-computer communication for physically challenged people. 

This work proposes a novel Hand Gesture Classification (HGC) model by employing a convolutional neural network based VGG-16 net \cite{ha2018} model by involving following steps:
\begin{enumerate}
\item Transfer learning – VGG-16 net is loaded with the weights obtained by training on ImageNet dataset \cite{im2015}, a standard dataset consisting of 1000 classes.
\item Partial training – The last few convolutional layers are set to be trained while others are freezed with preloaded weights.
\item Data Augmentation – The NUS dataset \cite{ku2010} is augmented first before training using Keras ImageDataGenerator tool \cite{fc2020}. Augmentations involve horizontal flip, horizontal image shift, vertical image shift, rotation of image.
\end{enumerate}

Section ~\ref{ls} briefly discusses on the work carried out in this area, while an introduction to the VGG-16 net is presented in Section ~\ref{vgg}. The suggested HGCM is extensively demonstrated in Section ~\ref{pa}, while the results are illustrated in Section ~\ref{exp}. At last, an abbreviated version of the work done is featured in Section ~\ref{conc}.

\section{Literature Survey}\label{ls}
This section introduces the reader on the considerable work done in the hand gesture classification and hand pose estimation during the early phases. A gesture detection mechanism has been developed in \cite{ba2015} that captures the hand points in 3D and uses machine learning for classification by producing an accuracy level of 62\%. However, the machine learning model approaches were later discarded due to their high loss level \cite{no2019}.

The authors use a sequence of convolutional layers for body pose estimation in \cite{si2014}. These layers produce a belief map of the poses that is refined in each subsequent layer. The authors in \cite{gh2015} uses the localized contour sequence as features being fed to Support Vector Machine (SVM) for classification of static hand gestures. Another paper \cite{pr2013} performs segmentation of hand from forearm region and classification using minimum distance classifier after geometry-based normalization. A hand pose estimation by generating the heat map of hand keypoints along with the use of decision forest is presented in \cite{ke2012}. Another work uses the ZIK and HOG for extracting hand point features and classifies using a multi layered Randomized Decision Forest (RDF) \cite{go2019}.

\section{Convolutional Neural Network based VGG-16 network}\label{vgg}
A deep Convolutional Neural Network (CNN) developed by VGG group of University of Oxford, known as VGG-16 net, is a collection of numerous convolutional and Max Pooling layers \cite{ha2018}. These layers, comprising of a set of independent filters, performs dot product with the input image to produce feature maps that are subsequently fed to the consecutive convolutional layer. The feature maps generated at each layer represents different features of the image.

During the training, the randomly initialized filters become the parameters which are learned automatically by the network. Max Pooling layers are added after the convolutional layers, specifically after introducing non-linearity to feature maps (for e.g. ReLU). Features maps generated from convolution operation holds the exact coordinates of features of the image and hence small movements of image like rotation, cropping, and shifting, etc., will create a different feature map rendering it to be a different image. Max Pooling layer solves this issue by operating over a feature map and creating a new feature map with reduced dimensions, consisting of maximum value from each patch of the previous feature map. The VGG-16 net contains one input layer \emph{input\_1}, one output layer and some blocks of computational layers like, \emph{block1}, \emph{block2}, \emph{block3}, \emph{block4}, \emph{block5} respectively. Each block comprises of few convolutional and max pooling layers to perform the convolution and max pooling operations. 

The Eq. (\ref{eq1}) and Eq. (\ref{eq2}) depict the computation of gradients required for the VGG-16 net model.
\begin{equation}\label{eq1}
E[g^2]_t = \beta E[g^2]_{t-1} + (1 -\beta)(\frac{\delta C}{\delta w})^2
\end{equation}

\begin{equation}\label{eq2}
w_t = w_{t-1} - \frac{\eta}{\sqrt{E[g^2]_t}}\frac{\delta C}{\delta w}
\end{equation}
where, $E[g]$ is the moving average of squared gradients, $\frac{\delta C}{\delta w}$ presents the gradient of the cost function corresponding to the weight $w$ at time $t$. $\beta$ denotes the moving average index having default value = 0.9 and $\eta$ refers the learning rate.

\section{Proposed System}\label{pa}
\begin{figure}
\centering
\includegraphics[width=\textwidth]{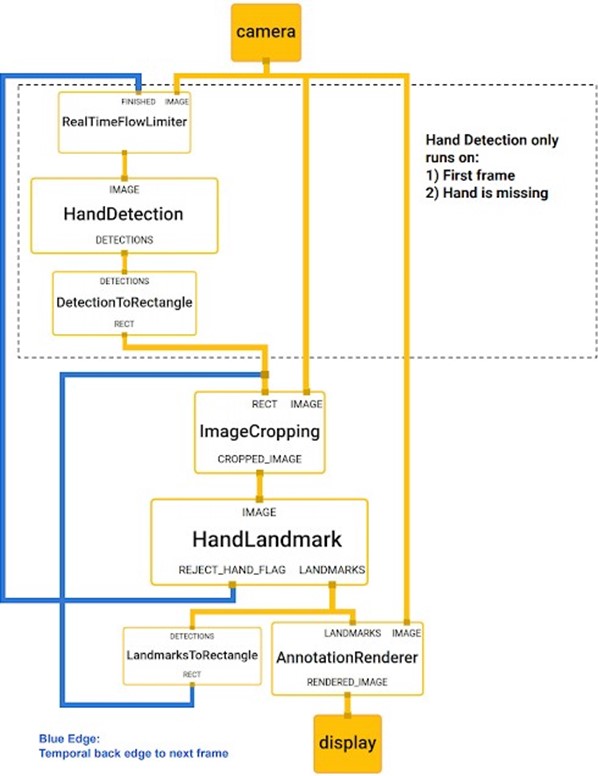}
\caption{Hand Gesture Classification Model} \label{fig3}
\end{figure}

The proposed Hand Gestures Classification (HGC) model employs the VGG-16 network for hand gestures classification that is pre-trained on ImageNet dataset \cite{im2015,fc2020}. This network is capable of learning 1000 different objects present in the training dataset. For validating the model, the NUS dataset consisting of 10 classes of hand gestures representing the alphabets `a' to `j' has been used \cite{ku2010}. The validated model then recognizes various hand gestures after being fed on a testing dataset, which comprises of data from various hand joints movements. This joint detection is done by MediaPipe, an open source framework for building pipelines to process perceptual data of AVs (audios/videos) \cite{lu2019}. Current methods of hand joint detection use desktop environments for inference and can achieve real time performance even on a mobile phone. It uses machine learning to extract the keypoints of a hand in a single frame for high fidelity hand and finger tracking as shown in Fig. \ref{fig3}. The hand tracking uses following approaches:
\begin{enumerate}
\item Hand bounding box palm detector - The palm detector pipeline first operates on the full image of the hand and returns an oriented hand bounding box presenting only the palm area. Initial hand locations are detected using a single short detector model.  
\item Key handpoints - Another model then operates on the cropped image and returns some key handpoints. This model is called hand landmark model that extracts the precise hand knuckle coordinates by applying regression (direct coordinate prediction) to the bounded hand image.
\item Classification - Finally, those data points form the testing dataset and given to the trained VGG-16 model as input for final gesture recognition.
\end{enumerate}

\section{Results and Discussions}\label{exp}
The proposed system is trained, validated, and tested in the MediaPipe framework based on Python and Keras libraries available in the Google Collab environment \cite{fc2020} on a Windows 10 operating system. Initially, the Google Collab environment loads the weights of the VGG-16 net model and the ImageNet dataset \cite{im2015} for training. During this time, the images are augmented in different combinations for accurate recognizance of various hand keypoints by capturing various knuckle positions and palm areas such as, up, down, forward, backward, etc. 

\begin{figure}
\centering
\begin{subfigure}{1\textwidth}
\centering
\includegraphics[width=\textwidth]{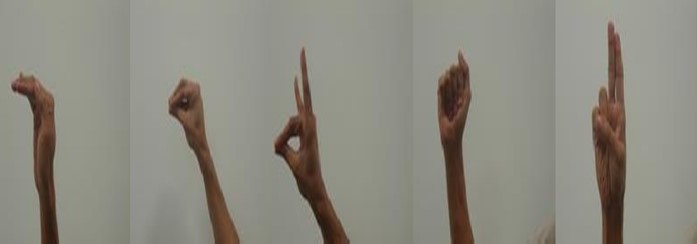}
\caption{} \label{fig5a}
\end{subfigure}
\begin{subfigure}{1\textwidth}
\centering
\includegraphics[width=\textwidth]{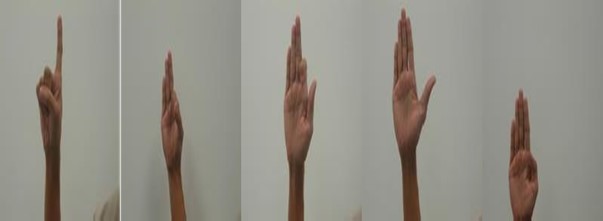}
\caption{} \label{fig5b}
\end{subfigure}
\caption{Various Inputs depicting the Alphabets (`a' - `j')} 
\label{fig5}
\end{figure}

First, the number of output classes is changed to 10 for validation purpose. The parameter values of each layer in the VGG-16 net, mentioned in Section \ref{vgg}, are present in \cite{fc2020}. The two fully connected (dense) layers are modified to have a dropout of 0.5 to avoid over fitting. The layers from input\_1 to block4\_pool are freezed and the layers from block5\_conv1 are set for training. The loss monitored is categorical \emph{cross entropy} and optimizer used is \emph{RMSprop} with a learning rate of 0.00002. Data augmentation like horizontal flip, angular rotations are carried out by using the Keras ImageDataGenerator \cite{fc2020} for reducing the loss and increasing generalization. The number of epochs is set to 30. For validation, the NUS dataset is used for strengthening the trained model as it contains 10 classes of hand gestures representing the alphabets `a' to `j' \cite{ku2010}. Afterwards, our own dataset has been created by using the Google's open source API MediaPipe \cite{fc2020}. Here, 10 different hand gestures representing the alphabets `a' to `j' are placed in front of the camera for capturing the key handpoints as presented in Fig. \ref{fig5}.

\begin{figure}
\centering
\includegraphics[width=\textwidth] {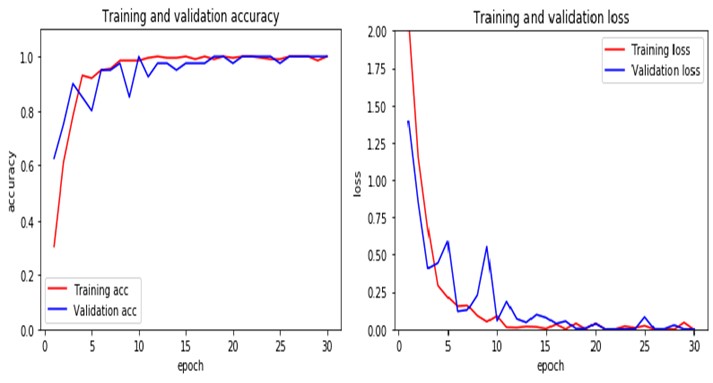}
\caption{Accuracy and Loss Graph of the Model}
\label{fig6}
\end{figure}

The performance of the model is evaluated by using \emph{Accuracy}, expressed in Eq. (\ref{eq3}), that measures the correctness of a classifier. 
\begin{equation}\label{eq3}
Accuarcy = \frac{TP + TN}{TP + TN + FP + FN}
\end{equation}
where, FP, TN, TP and FN refers to the false positives, true negatives, true positives and false positives respectively. Furthermore, another measuring index known as \emph{Cross Entropy Loss}, mentioned in Eq. (\ref{eq4}), is used to compute the error rate/difference between the predicted and actual output. 
\begin{equation}\label{eq4}
Loss = - \frac{1}{N}\sum_{i=1}^{N}{y_i \cdot log(\hat{y}_i)} + (1-y_i) \cdot log(1-\hat{y}_i)
\end{equation}
where, $y$ and $\hat{y}$ denote the target and predicted outputs respectively, while $N$ refers to the output size. Fig. \ref{fig6} demonstrates the accuracy and loss of the model during the training and validation phase. The proposed hand gesture classification model yields 98.33\% accuracy by recognizing the inputted images belonging to the 10 classes.

\section{Conclusions}\label{conc}
In this work, a customized convolution neural network, known as VGG-16 net, is used to accurately recognize various hand gestures. Initially, the VGG-16 net is trained on the ImageNet dataset in the Google Collab environment to produce the trained model. Various combinations of training parameters are used for tuning the performance of the trained model. Furthermore, the NUS dataset, consisting of 10 classes representing the alphabets `a' to `j', has been used for validating the trained model. Besides, 10 hand gestures are captured using the MediaPipe framework for final classification. The model has been able to identify the hand gestures successfully with 98.33\% accuracy.

\end{document}